%% file: main.tex
\documentclass[english,conference]{IEEEtran}
\usepackage[T1]{fontenc}
\usepackage[latin9]{inputenc}
\usepackage{array}
\usepackage{multirow}
\usepackage{amsmath}
\usepackage{amssymb}
\usepackage{graphicx}
\usepackage{microtype}

\makeatletter

\providecommand{\tabularnewline}{\\}

\usepackage{times}
\usepackage{epsfig}
\usepackage{graphicx}
\usepackage{flushend}

\usepackage{microtype}

\makeatother

\usepackage{babel}
\begin{document}
\title{Neural Reasoning, Fast and Slow, for Video Question Answering}
\author{Thao Minh Le, Vuong Le, Svetha Venkatesh, Truyen Tran\\
Applied Artificial Intelligence Institute, Deakin University, Australia\\
 \texttt{\small{}\{lethao,vuong.le,svetha.venkatesh,truyen.tran\}@deakin.edu.au}}
\maketitle
\begin{abstract}
\input{abs.tex}
\end{abstract}

\section{Introduction}

\input{intro.tex}

\section{Related Work \label{sec:Related-Work}}

\input{related.tex}

\section{Method \label{sec:Method}}

\input{method.tex}

\section{Experiments \label{sec:Experiments}}

\input{exps.tex}

\section{Discussion \label{sec:Discussion}}

\input{discussion.tex}

{\small{}\bibliographystyle{plain}
\bibliography{egbib,ME,truyen}
}{\small\par}
\end{document}

%% file: abs.tex
What does it take to design a machine that learns to answer natural
questions about a video? A Video QA system must simultaneously understand
language, represent visual content over space-time, and iteratively
transform these representations in response to lingual content in
the query, and finally arriving at a sensible answer. While recent
advances in lingual and visual question answering have enabled sophisticated
representations and neural reasoning mechanisms, major challenges
in Video QA remain on dynamic grounding of concepts, relations and
actions to support the reasoning process. Inspired by the dual-process
account of human reasoning, we design a \emph{dual process neural
architecture}, which is composed of a question-guided video processing
module (System 1, fast and reactive) followed by a generic reasoning
module (System 2, slow and deliberative). System 1 is a hierarchical
model that encodes visual patterns about objects, actions and relations
in space-time given the textual cues from the question. The encoded
representation is a set of high-level visual features, which are then
passed to System 2. Here multi-step inference follows to iteratively
chain visual elements as instructed by the textual elements. The system
is evaluated on the SVQA (synthetic) and TGIF-QA datasets (real),
demonstrating competitive results, with a large margin in the case
of multi-step reasoning. 

%% file: intro.tex
A long standing goal in AI is to design a learning machine capable
of reasoning about dynamic scenes it sees. A powerful demonstration
of such a capability is answering unseen natural questions about a
video. Video QA systems must be able to learn, acquire and manipulate
visual knowledge distributed through space-time conditioned on the
compositional linguistic cues. Recent successes in image QA \cite{anderson2018bottom,hu2017learning,hudson2019learning,yi2018neural}
pave possible roads, but Video QA is largely under-explored \cite{song2018explore,li2019beyond}.
Compared to static images, video poses new challenges, primarily due
to the inherent dynamic nature of visual content over time \cite{gao2018motion,wang2018movie}.
At the lowest level, we have correlated motion and appearance \cite{gao2018motion}.
At higher levels, we have objects that are persistent over time, actions
that are local in time, and the relations that can span over an extended
length.

Searching for an answer from a video facilitates solving interwoven
sub-tasks in both the visual and lingual spaces, probably in an iterative
and compositional fashion. In the visual space, the sub-tasks at each
step involve extracting and attending to objects, actions, and relations
in time and space. In the textual space, the tasks involve extracting
and attending to concepts in the context of sentence semantics. A
plausible approach to Video QA is to prepare video content to accommodate
the retrieval of information specified in the question \cite{jang2017tgif,kim2017deepstory,zeng2017leveraging}.
But this has not yet offered a more complex reasoning capability that
involves multi-step inference and handling of compositionality. More
recent works have attempted to add limited reasoning capability into
the system through memory and attention mechanisms that are tightly
entangled with visual representation \cite{gao2018motion,li2019beyond}.
These systems are thus non-modular, and less comprehensible as a result.

Our approach to Video QA is to disentangle the processes of visual
pattern recognition\emph{ }and\emph{ }compositional reasoning \cite{yi2018neural}.
This division of labor realizes a \emph{dual process} cognitive view
that the two processes are qualitatively different: visual cognition
can be reactive and domain-specific but reasoning is usually deliberative
and domain-general \cite{evans2008dual,kahneman2011thinking}. In
our system, pattern recognition precedes and makes its output accessible
to the reasoning process. More specifically, at the visual understanding
level, we derive a hierarchical model over time, dubbed Clip-based
Relational Network (CRN), that can accommodate objects, actions, and
relations in space-time. This is followed by a generic differentiable
reasoning module, known as Memory-Attention-Composition (MAC) network
\cite{hudson2018compositional}, which iteratively manipulates a set
of objects in the knowledge base given a set of cues in the query,
one step at a time. In our setting, MAC takes prepared visual content
as a knowledge base, and iteratively co-attends to the textual concepts
and the visual concepts/relations to extract the answer. The overall
dual-process system is modular and fully differentiable, making it
easy to compose modules and train.

\begin{figure*}
\begin{centering}
\includegraphics[width=0.95\textwidth]{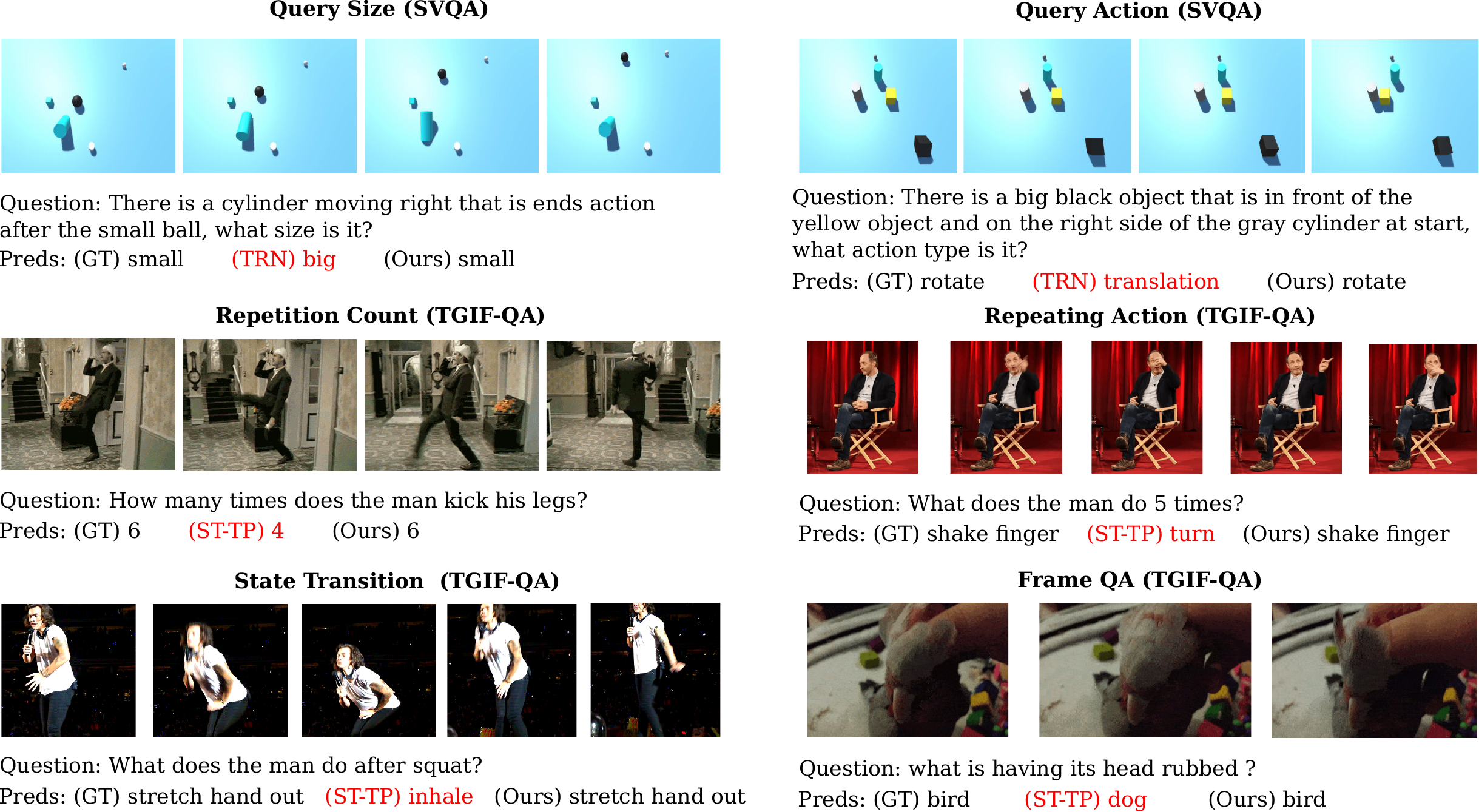}
\par\end{centering}
\caption{Examples of SVQA and TGIF-QA dataset. GT: ground truth; TRN: our baseline
utilizing TRN \cite{zhou2018temporal}; ST-TP: method introduced in
\cite{jang2017tgif}. Best viewed in color. \label{fig:qualcomp}}
\end{figure*}

We validate our dual process model on two large public datasets, the
TGIF-QA and the SVQA. The TGIF-QA is a real dataset, and is relatively
well-studied \cite{gao2018motion,jang2017tgif,li2019beyond}. See
Fig.~\ref{fig:qualcomp}, last two rows for example frames and question
types. The SVQA is a new synthetic dataset designed to mitigate the
inherent biases in the real datasets and to promote multi-step reasoning
\cite{song2018explore}. Several cases of complex, multi-part questions
are shown in Fig.~\ref{fig:qualcomp}, first row. On both datasets,
the proposed model (CRN+MAC) achieves new records, and the margin
on the SVQA is qualitatively different from the best known results.
Some example responses are displayed in Fig.~\ref{fig:qualcomp},
demonstrating how our proposed method works in different scenarios.

Our contributions are 1) Introducing a modular neural architecture
for learning to reason in Video QA. The system implements dual process
theory by disentangling reactive visual cognition and language understanding
from deliberative reasoning. 2) Proposing a hierarchical model for
preparing video representation taking into account of query-driven
frame selectivity within a clip and temporal relations between clips.

%% file: related.tex
\paragraph*{Video representation in Video QA}

Available methods for Video QA typically relied on recurrent networks
or 3D convolutions to extract video features. Variations of LSTM were
used in \cite{kim2017deepstory} with a bidirectional LSTM, \cite{zeng2017leveraging}
in the form of a two-staged LSTM. Likewise, \cite{gao2018motion,li2019beyond}
used two levels of GRU, the first one for extracting ``facts'' and
the second one in each iteration of the memory based reasoning. In
another direction, convolutional networks were used to integrate visual
information with either 2D or 3D kernels \cite{jang2017tgif,li2019beyond}.

Different from these two traditional trends, in this work we propose
CRN, a query-driven hierarchical relational feature extraction strategy,
which supports strong modeling for both near-term and far-term spatio-temporal
relations. CRN supports multiple levels of granularity in the temporal
scale. This development is necessary to address the nondeterministic
queries in Video QA tasks.

\paragraph*{Neural reasoning for Video QA}

The work in \cite{kim2017deepstory,zeng2017leveraging} both utilized
memory network as a platform to retrieve the information in the video
features related to the question embedding. More recent Video QA
methods started interleaving simple reasoning mechanisms into the
pattern matching network operations. In \cite{jang2017tgif}, Jang
\emph{et al.} calculated the attention weights on the video LSTM features
queried by the question.  In an effort toward deeper reasoning, Gao
\emph{et al.} \cite{gao2018motion} proposed to parse the two-stream
video features through a dynamic co-memory module which iteratively
refines the episodic memory. Lately \cite{li2019beyond} used self-attention
mechanisms to internally contemplate about video and question first,
then put them through a co-attention block to match the information
contained in the two sources of data. For complex structured videos
with multimodal features, recent method leveraged memories \cite{fan2019heterogeneous,kim2017deepstory,na2017read}
to store multimodal features into episodic memory for later retrieval
of related information to the question. More sophisticated reasoning
mechanisms are also developed with hierarchical attention \cite{liang2018focal},
multi-head attention \cite{kim2018multimodal} or multi-step progressive
attention memory \cite{kim2019progressive} to jointly reason on video/audio/text
concurrent signals.

The current trend set by these works pushes the sophistication of
the reasoning processes on finding the correlation between data pattern
and the query. Pattern recognition and reasoning are tightly entangled,
and reasoning tends to be specific to visual/textual patterns as a
result. Compared to the previous works, our framework steps toward
disentangling these two processes. 

\paragraph*{Dual process systems}

Reasoning systems that exhibit behaviors consistent with dual process
theories are typically neural-symbolic hybrids (e.g., see \cite{garcez2019neural}
for an overview). In \cite{yi2018neural}, visual pattern recognition
modules form elements of a symbolic program whose execution would
find answers for image question answering. Different from \cite{yi2018neural},
we rely on implicit reasoning capability in a fully differentiable
neural system \cite{bottou2014machine,hudson2018compositional}.

%% file: method.tex
\begin{figure*}
\centering{}\includegraphics[width=0.95\textwidth]{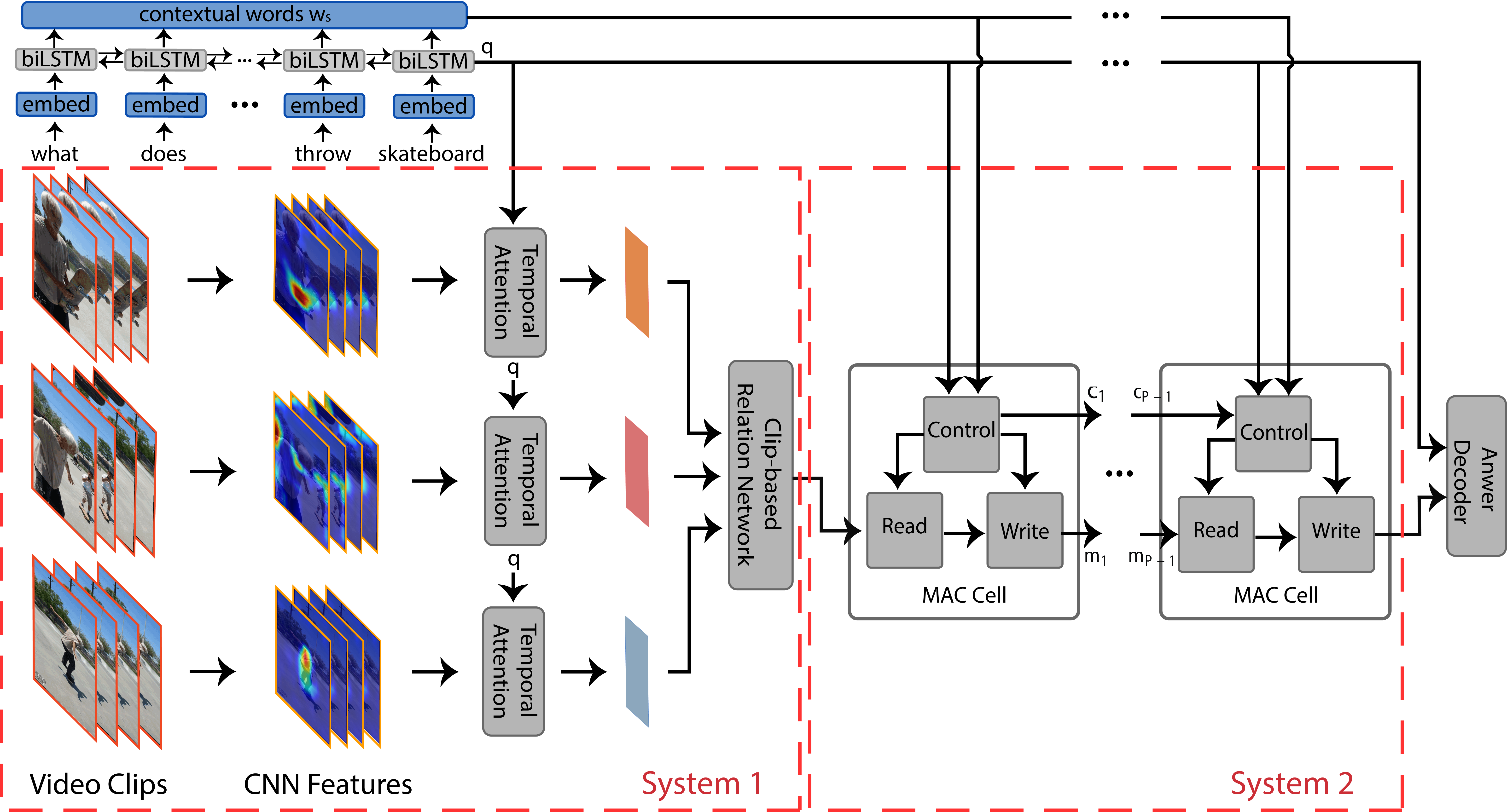}\caption{Overview of Network Architecture for Video QA. The model is viewed
as a dual process system of hierarchical video representation with
Clip-based Relation Network (CRN) and high-level multi-step reasoning
with MAC cells, in which the two processes are guided by textual cues.
Inputs of CRN are the aggregated features of equal-size clips obtained
by a temporal attention mechanism. The high-level reasoning module
iteratively co-attends to the contextual words of a given question
and the visual concepts/relations prepared by the CRN unit to extract
relevant visual information to the answer. At the end of the network,
an answer decoder taking as input the joint representation of the
question feature and the retrieved visual information is used for
prediction. \label{fig:architecture}}
\end{figure*}

In this section, we present our main contribution to addressing the
challenges posed in Video QA. In particular, we propose a modular
end-to-end neural architecture, as illustrated in Fig.~\ref{fig:architecture}.

\subsection{Dual Process System View}

Our architecture is partly inspired by the \emph{dual process theory}
dictating that there are two loosely coupled cognitive processes serving
separate purposes in reasoning: the lower pattern recognition that
tends to be associative, and the higher-order reasoning faculty that
tends to be deliberative \cite{evans2008dual,kahneman2011thinking}.
Translated into our Video QA scenarios, we have the pattern recognition
process for extracting visual features, representing objects and relations,
and making the representation accessible to the higher reasoning process.
The interesting and challenging aspects come from two sources. First,
video spans over both space and time, and hence calling for methods
to deal with object persistence, action span and repetition, and long-range
relations. Second, Video QA aims to respond to the textual query,
hence the two processes should be conditional, that is, the textual
cues will guide both the video representation and reasoning.

For language coding, we make use of the standard bi-LSTM with GloVe
word embeddings. Let $S$ be a given question's length, we subsequently
obtain two sets of linguistic clues: contextual words $\left\{ w_{s}|w_{s}\in\mathbb{R}^{d}\right\} _{s=1}^{S}$
which are the output states of the LSTM at each step, and the question
vector $q=[\overleftarrow{w_{1}};\overrightarrow{w_{S}}],q\in\mathbb{R^{\text{d}}}$
which is the joint representation of the final hidden states from
forward and backward LSTM passes.

 We treat a video as a composition of video clips, in which each
clip can be viewed as an activity. While previous studies have explored
the importance of hierarchical representation of video \cite{zhu2018end},
we hypothesize that it is also vital to model the relationships between
clips. Inspired by \cite{santoro2017simple} and a recent work \cite{zhou2018temporal}
on action recognition, known as Temporal Relation Network (TRN), we
propose a Clip-based Relation Network (CRN) for video representation,
where clip features are selectively query-dependent. It is expected
that CRN is more effective in terms of modeling a temporal sequence
than the simplistic TRN which comes with a certain number of sampled
frames. The CRN represents the video as a tensor for the use in the
reasoning stage.

The reasoning process, due to its deliberative nature, involves multiple
steps in an iterative fashion. We utilize Memory-Attention-Composition
(MAC) cells \cite{hudson2018compositional} for the task due to its
generality and modularity. More specially, the MAC cells are called
repeatedly conditioned on the textual cues to manipulate information
from given video representations as a knowledge base. Finally, information
prepared by the MAC, combined with the textual cues is presented to
a decoder to produce an answer.

In short, our system consists of three components where the outputs
of one component are the inputs to another: (1) temporal relational
pattern recognition, (2) multi-step reasoning with MAC cells and (3)
answer decoders. We detail these components in what follows.

\subsection{Temporal Relational Pattern Recognition \label{subsec:CRN}}

\begin{figure}
\centering{}\includegraphics[width=1\columnwidth]{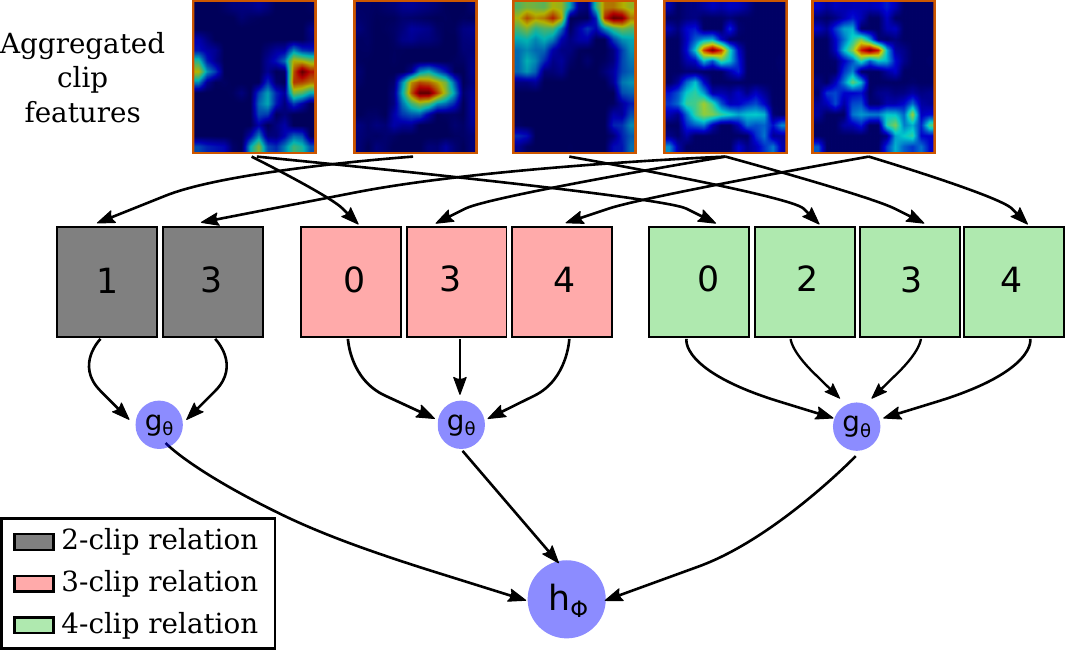}\caption{Illustration of Clip-based Relation Network (CRN). Aggregated features
of equal size clips are fed into k-clip relation modules. Inputs to
relation modules are selected on a random basis whilst keeping their
temporal order unchanged. In this figure, our CRN represents a video
as aggregated features of five video clips using 2-clip relation,
3-clip relation, and 4-clip relation modules. This results in the
final feature of the same shape as one clip feature.\label{fig:crn}}
\end{figure}
Given a video of continuous frames, we begin with dividing the video
into $L$ equal-length clips $C=\left(C_{1},...,C_{l},...,C_{L}\right)$.
Each clip $C_{l}$ of $T$ frames is represented by $C_{l}=\{V_{l,t}\mid V_{l,t}\in\mathbb{R}^{W\times H\times D}\}_{t=1}^{T}$,
where $V_{l,t}$ is frame features extracted by ResNet-101 \cite{he2016deep}
of the $t$-th frame in clip $C_{l}$; $W,H,D$ are dimensions of
the extracted features. Frame-level features are subsequently projected
to a $d$ dimensional space via linear transformations, resulting
$C_{l}=\{V_{l,t}^{\prime}\mid V_{l,t}^{\prime}\in\mathbb{R}^{W\times H\times d}\}_{t=1}^{T}$.

As consecutive frames are redundant or irrelevant to the question,
it is crucial to selectively attend to frames. In addition, this would
greatly reduce the computational cost. We thus utilize a temporal
attention mechanism conditioned on the question vector $q$ to compute
the aggregated clip feature $\hat{C}_{l}$ of the corresponding clip
$C_{l}$:

\begin{align}
V_{l,t}^{\text{pool}} & =\frac{1}{W\times H}\sum_{w=1}^{W}\sum_{h=1}^{H}V_{l,t,w,h}^{\prime};V_{l,t}^{\text{pool}}\in\mathbb{R}^{d},\\
s_{l,t} & =W\left(\left(W^{q}q+b^{q}\right)\odot\left(W^{v}V_{l,t}^{\text{pool}}+b^{v}\right)\right),\\
\hat{C_{l}} & =\sum_{t=1}^{T}V_{l,t}^{\prime}\cdot\textrm{softmax}(s_{l,t}),
\end{align}
where, $W,W^{q},W^{v},b^{q}$ and $b^{v}$ are learnable weights,
and $\odot$ is element-wise multiplication.

To account for relations between clips, we borrow the strategy of
TRN described in \cite{zhou2018temporal} which adapts and generalizes
the proposal in \cite{santoro2017simple} to temporal domain. Different
from \cite{zhou2018temporal}, our relational network operates at
the clip level instead of frame level. More specifically, the $k$-order
relational representation of video is given as

\begin{equation}
R^{(k)}\left(C\right)=h{}_{\Phi}\left(\sum_{l_{1}<l_{2}...<l_{k}}g{}_{\theta}\left(\hat{C}_{l_{1}},\hat{C}_{l_{2}},...,\hat{C}_{l_{k}}\right)\right),\label{eq:k-level-relation}
\end{equation}
for $k=2,3,..,K$, where $h_{\phi}$ and $g_{\theta}$ are any aggregation
function with parameters $\phi$ and $\theta$, respectively. We term
this resulted model as \emph{Clip-based Relation Network} (CRN). Fig.~\ref{fig:crn}
illustrates our procedure for our CRN.

\paragraph{Remark}

The CRN subsumes TRN as a special case when $T\rightarrow1$. However,
by computing the relations at the clip level, we can better model
the hierarchical structure of video and avoid computational complexity
inherent in TRN. For example, we neither need to apply sparse sampling
of frames nor use the multi-resolution trick as in TRN. Consider a
lengthy video sequence, in TRN, the chance of having pairs of distantly
related frames is high, hence, their relations are less important
than those of near-term frames. In the worst case scenario, those
pairs can become noise to the feature representation and damage the
reasoning process in later stages. In contrast, not only our clips
representation can preserve such near-term relations but also the
far-term relations between short snippets of a video can be guaranteed
with the CRN.

\subsection{Multi-step Centralized Reasoning}

Higher-order reasoning on the rich relational temporal patterns is
the key for reliably answering questions.  Our approach is to disentangle
the slow, deliberative reasoning steps out of fast, automatic feature
extraction and temporal relation modeling. This ``slow-thinking''
reasoning is done with a dedicated module that iteratively distills
and purifies the key relational information contained in the CRN features.

In our experiments, we use the MAC network \cite{hudson2018compositional}
as the option for the reasoning module. At the core of MAC network
are the recurrent cells called \emph{control units}, collaborating
with \emph{read units} and \emph{write units} to iteratively make
reasoning operations on a knowledge base using a sequence of clues
extracted from the query. Compared to mixed-up feature extraction/reasoning
mechanisms, the control units give the MAC network distinctive features
of a centralized reasoning module that can make a well-informed decision
on attention and memory reads/writes. MAC is also powered by the flexible
retrieving/processing/reference mechanism while processing the query
and looking up in the knowledge base. These characteristics are well
suited to explore the rich, condensed relational information in CRN
features. The iterative reasoning process of MAC supports a level
of error self-correcting ability that also helps to deal with the
possible remaining redundancy and distraction.

In our setup, the knowledge base $B$ used in MAC network is gathered
from the CRN features from all available orders:
\begin{equation}
B=\sum_{k=2}^{K}R^{(k)}\left(C\right),
\end{equation}
where, $R^{(k)}\left(C\right)$ are the $k$-order CRN representations
calculated as in Eq.~(\ref{eq:k-level-relation}).

For each reasoning step $i$, the relevant aspects of question to
this step are estimated from $q$:

\begin{equation}
q_{i}=W_{i}^{q}q+b_{i}^{q},
\end{equation}
where, $W_{i}^{q}$ and $b_{i}^{q}$ are network weights.

Let $[;]$ denote the concatenation operator of two tensors. Based
on the pair of clues contextual words and step-aware question vector
($\{w_{s}\}_{s=1}^{S},q_{i})$, recall that $S$ is a given question's
length, and the control state of the previous reasoning step $c_{i-1}$,
the control unit calculates a soft self-attention weight $\alpha_{i,s}^{\text{control}}$
over words in the question:
\begin{equation}
F_{i}=W_{i}^{c1}[W_{i}^{c0}c_{i-1};q_{i}],
\end{equation}

\begin{equation}
\alpha_{i,s}^{\text{control}}=\textrm{softmax}(W_{i}^{\alpha}(F_{i}\odot w_{s})+b^{\alpha}),
\end{equation}
and infers the control state at this step:
\begin{equation}
c_{i}=\sum_{s=1}^{S}\alpha_{i,s}^{\text{control}}w_{s}.
\end{equation}

The read unit uses this control signal and the prior memory $m_{i-1}$
to calculate the read attention weights $\alpha_{i,x,y}^{\text{read}}$
for each location $x,y$ in the knowledge base $B$ and retrieves
the related information:
\begin{equation}
r_{i}=\sum_{x,y}\alpha_{i,x,y}^{\text{read}}B_{x,y},
\end{equation}
where,

\begin{equation}
I_{i,x,y}=[m_{i-1}\odot B_{x,y};B_{x,y}],
\end{equation}
\begin{equation}
I_{i,x,y}^{\prime}=W_{i}^{I}I_{i,x,y},
\end{equation}

\begin{equation}
\alpha_{i,x,y}^{\text{read}}=\textrm{softmax}(W_{i}^{\alpha}(c_{i}\odot I_{i,x,y}^{\prime})+b^{\alpha}).
\end{equation}

To finish each reasoning iteration, the write unit calculates the
intermediate reasoning result $m_{i}$ by updating previous record
$m_{i-1}$ with the new information derived from the retrieved knowledge
$r_{i}$ , say $m_{i}=f(m_{i-1},r_{i})$. In our experiments, the
function $f$ is simply a linear transformation on top of a vector
concatenation operator.

At the end of the process of $P$ steps ($P$ MAC cells), the final
memory state $m_{P}$ emerges as the output of the reasoning module
used by the answer decoders in the next stage.

\subsection{Answer Decoders}

Following \cite{jang2017tgif,song2018explore}, we adopt different
answer decoders depending on the tasks. These include open-ended words,
open-ended number, and multi-choice question.

For open-ended words (e.g. those in Frame QA in TGIF-QA dataset and
all QA pairs in SVQA dataset \textendash{} see Sec.~\ref{subsec:Datasets}),
we treat them as multi-label classification problems of $\mathbb{\mathcal{V}}$
labels defined in an answer space $\mathcal{A}$. Hence, we employ
a classifier which composes 2-fully connected layers, and the following
softmax, and takes as of input the combination of the memory state
$m_{P}$ and the question representation $q$:

\begin{equation}
p=\textrm{softmax}(W^{o2}(W^{o1}[m_{p};W^{q}q+b^{q}]+b^{o1})+b^{o2}),
\end{equation}
where, $p\in\mathbb{R^{\mathcal{V}}}$ is a confidence vector of probabilities
of labels. The cross-entropy is used as the loss function of the network
in this case.

Similarly, we also use a linear regression function to predict real-value
numbers (repetition count) directly from the joint representation
of $m_{P}$ and $q$. We further pass the regression output through
a rounding function for prediction:

\begin{equation}
s=\lfloor W^{o2}(W^{o1}[m_{p};W^{q}q+b^{q}]+b^{o1})+b^{o2}\rfloor.
\end{equation}
Mean Squared Error (MSE) is used as the loss function during the training
process in this case.

Regarding the multi-choice question type, which includes repeating
action and state transition in TGIF-QA dataset in later experiments,
we treat each answer candidate of a short sentence in the same way
as with questions. In particular, we reuse one MAC network for both
questions and answer candidates in which network parameters are shared.
As a result, there are two types of memory output, one derived by
question $m_{q,P}$, and the other one conditions on answer candidates
$m_{a,P}$. Inputs to a classifier are from four sources, including
$m_{q,P}$, $m_{a,P}$ , question representation $q$ and answer candidates
$a$:

\begin{equation}
y=[m_{q,P};m_{a,P};W^{q}q+b^{q};W^{a}a+b^{a}],
\end{equation}

\begin{equation}
y^{\prime}=\sigma(W^{y}y+b^{y});\sigma=\text{ELU(.)},
\end{equation}
Finally, a linear regression is used to output an answer index:

\begin{equation}
s=W^{y^{\prime}}y^{\prime}+b^{y^{\prime}}.
\end{equation}
The model in this case is trained with hinge loss of pairwise comparisons,
$\text{max}\left(0,1+s^{n}-s^{p}\right)$, between scores for incorrect
$s^{n}$ and correct answers $s^{p}$.

%% file: exps.tex
\subsection{Datasets \label{subsec:Datasets}}

We evaluate our proposed method on two recent public datasets: Synthetic
Video Question Answering (SVQA) \cite{song2018explore} and TGIF-QA
\cite{jang2017tgif}.

\paragraph{SVQA}

This dataset is a benchmark for multi-step reasoning. Resembling the
CLEVR dataset \cite{johnson2017clevr} for traditional visual question
answering task, SVQA provides long questions with logical structures
along with spatial and temporal interactions between objects. SVQA
was designed to mitigate several drawbacks of current Video QA datasets
including language bias and the shortage of compositional logical
structure in questions. It contains 120K QA pairs generated from 12K
videos that cover a number of question types such as count, exist,
object attributes comparison and query.

\paragraph{TGIF-QA}

This is currently the largest dataset for Video QA, containing 165K
QA pairs collected from 72K animated GIFs. This dataset covers four
sub-tasks mostly to address the unique properties of video including
repetition count, repeating action, state transition and frame QA.
Of the four tasks, the first three require strong spatio-temporal
reasoning abilities. \emph{Repetition Count}: This is one of the most
challenging tasks in Video QA where machines are asked count the repetitions
of an action. For example, one has to answer questions like ``How
many times does the woman shake hips?''. This is defined as an open-ended
task with 11 possible answers in total ranging from 0 to 10+. \emph{Repeating
Action}: This is a multiple choice task of five answer candidates
corresponding to one question. The task is to identify the action
that is repeated for a given number of times in the video (e.g. ``what
does the dog do 4 times?''). \emph{State Transition}: This is also
a multiple choice task asking machines to perceive the transition
between two states/events. There are certain states characterized
in the dataset including facial expressions, actions, places and object
properties. Questions like ``What does the woman do before turn
to the right side?'' and ``What does the woman do after look left
side?'' aim at identifying previous state and next state, respectively.
\emph{Frame QA}:  This task is akin to the traditional visual QA
where the answer to a question can be found in one of the frames in
a video. None of temporal relations is necessary to answer questions.

\subsection{Implementation Details \label{subsec:Implementation-Details}}

Each video is segmented into five equal clips, each of which has eight
consecutive frames. The middle frame of each clip is determined based
on the length of the video. We take \emph{conv4 }output features from
ResNet-101 \cite{he2016deep} pretrained on ImageNet as the visual
features of each video frame. Each frame feature has dimensions of
$14\times14\times1024$. Each word in questions and answer candidates
in case of multiple choice question is embedded into a vector of dimension
300 and initialized by pre-trained GloVe embeddings \cite{pennington2014glove}.
 Unless otherwise stated, we use $P=12$ MAC cells for multi-step
reasoning in our network, similar to what described in \cite{hudson2018compositional},
while all hidden state sizes are set to $512$ for both CRN and MAC
cells.

Our network is trained using Adam, with a learning rate of $5\times10^{-5}$
for repetition count and $10^{-4}$ for other tasks, and a batch size
of 16. The SVQA is split into three parts with proportions of 70-10-20\%
for training, cross-validation, and testing set, accordingly. As for
TGIF-QA dataset, we take 10\% of training videos in each sub-task
as the validation sets. Reported results are at the epochs giving
best of performance on the validation sets. 

\paragraph{Evaluation Metrics}

For the TGIF-QA, to be consistent with prior works \cite{jang2017tgif,gao2018motion,li2019beyond},
we use accuracy as the evaluation metric for all tasks except the
repetition count task whose evaluation metric is Mean Square Error
(MSE). For the SVQA, we report accuracy for all sub-tasks, which are
considered as multi-label classification problems.

\subsection{Results \label{subsec:Results}}

\subsubsection{Ablation Studies}

\begin{table}
\caption{Ablation studies. ({*}) For count, the lower the better. \label{tab:ablation1}}

\centering{}{\footnotesize{}}%
\begin{tabular}{l|c|c|c|c|c}
\hline 
\multirow{2}{*}{{\footnotesize{}Model}} & \multirow{2}{*}{{\footnotesize{}SVQA}} & \multicolumn{4}{c}{{\footnotesize{}TGIF-QA ({*})}}\tabularnewline
\cline{3-6} \cline{4-6} \cline{5-6} \cline{6-6} 
 &  & {\footnotesize{}Action} & {\footnotesize{}Trans.} & {\footnotesize{}Frame} & {\footnotesize{}Count}\tabularnewline
\hline 
{\footnotesize{}Linguistic only} & {\footnotesize{}42.6} & {\footnotesize{}51.5} & {\footnotesize{}52.8} & {\footnotesize{}46.0} & {\footnotesize{}4.77}\tabularnewline
{\footnotesize{}Ling.+S.Frame} & {\footnotesize{}44.6} & {\footnotesize{}51.3} & {\footnotesize{}53.4} & {\footnotesize{}50.4} & {\footnotesize{}4.63}\tabularnewline
{\footnotesize{}S.Frame+MAC} & {\footnotesize{}58.1} & {\footnotesize{}67.8} & {\footnotesize{}76.1} & {\footnotesize{}57.1} & {\footnotesize{}4.41}\tabularnewline
{\footnotesize{}Avg.Pool+MAC} & {\footnotesize{}67.4} & {\footnotesize{}70.1} & {\footnotesize{}77.7} & {\footnotesize{}58.0} & {\footnotesize{}4.31}\tabularnewline
{\footnotesize{}TRN+MAC} & {\footnotesize{}70.8} & {\footnotesize{}69.0} & {\footnotesize{}78.4} & {\footnotesize{}58.7} & {\footnotesize{}4.33}\tabularnewline
{\footnotesize{}CRN+MLP} & {\footnotesize{}49.3} & {\footnotesize{}51.5} & {\footnotesize{}53.0} & {\footnotesize{}53.5} & {\footnotesize{}4.53}\tabularnewline
\hline 
\textbf{\footnotesize{}CRN+MAC} & \textbf{\footnotesize{}75.8} & \textbf{\footnotesize{}71.3} & \textbf{\footnotesize{}78.7} & \textbf{\footnotesize{}59.2} & \textbf{\footnotesize{}4.23}\tabularnewline
\hline 
\end{tabular}{\footnotesize\par}
\end{table}

\begin{table}
\caption{Ablation studies with different reasoning iterations. ({*}) For count,
the lower the better.\label{tab:ablation2}}

\centering{}{\small{}}%
\begin{tabular}{c|c|c|c|c}
\hline 
\multirow{2}{*}{{\footnotesize{}Reasoning iterations}} & \multicolumn{4}{c}{{\footnotesize{}TGIF-QA ({*})}}\tabularnewline
\cline{2-5} \cline{3-5} \cline{4-5} \cline{5-5} 
 & {\footnotesize{}Action} & {\footnotesize{}Trans.} & {\footnotesize{}Frame} & {\footnotesize{}Count}\tabularnewline
\hline 
{\footnotesize{}4} & {\footnotesize{}69.9} & {\footnotesize{}77.6} & {\footnotesize{}58.5} & {\footnotesize{}4.30}\tabularnewline
{\footnotesize{}8} & {\footnotesize{}70.8} & \textbf{\footnotesize{}78.8} & {\footnotesize{}58.6} & {\footnotesize{}4.29}\tabularnewline
{\footnotesize{}12} & \textbf{\footnotesize{}71.3} & {\footnotesize{}78.7} & \textbf{\footnotesize{}59.2} & \textbf{\footnotesize{}4.23}\tabularnewline
\hline 
\end{tabular}{\small\par}
\end{table}

To demonstrate the effectiveness of each component on the overall
performance of the proposed network, we first conduct a series of
ablation studies on both the SVQA and TGIF-QA datasets. The ablation
results are presented in Table~\ref{tab:ablation1}, \ref{tab:ablation2}
showing progressive improvements, which justify the added complexity.
We explain below the baselines.

\textbf{Linguistic only:} With this baseline, we aim to assess how
much linguistic information affects overall performance. From Table~\ref{tab:ablation1},
it is clear that TGIF-QA is highly linguistically biased while the
problem is mitigated with SVQA dataset to some extent.

\textbf{Ling.+S.Frame:} This is a very basic model of VQA that combines
the encoded question vector with CNN features of a random frame taken
from a given video. As expected, this baseline gives modest improvements
over the model using only linguistic features.

\textbf{S.Frame+MAC:} To demonstrate the significance of multi-step
reasoning in Video QA, we randomly select one video frame and then
use its CNN feature maps as the knowledge base of MAC. As the SVQA
dataset contains questions with compositional sequences, it greatly
benefits from performing reasoning process in a multi-step manner.

\textbf{Avg.Pool+MAC:} A baseline to assess the significance of temporal
information in the simplest form of average temporal pooling comparing
to ones using a single frame. We follow \cite{zhou2018temporal} to
sparely sample 8 frames which are the middle frames of the equal size
segments from a given video. As can be seen, this model is able to
achieve significant improvements comparing to the previous baselines
on both datasets. Due to the linguistic bias, the contribution of
visual information to the overall performance on the TGIF-QA is much
modest than that on the SVQA.

\textbf{TRN+MAC:} This baseline is a special case of ours where we
flatten the hierarchy, and the relation network is applied at the
frame level, similar to what proposed in \cite{zhou2018temporal}.
The model mitigates the limit of feature engineering process for video
representation of a single frame as well as simply temporal average
pooling. Apparently, using a single frame loses crucial temporal information
of the video and is likely to fail when strong temporal reasoning
capability plays a crucial role, particularly in state transition
and counting. We use visual features processed in the Avg.Pool+MAC
experiment to feed into a TRN module for fair comparisons.  TRN improves
by more than 12\% of overall performance from the one using a single
video frame on the SVQA, while the increase for state transition task
of the TGIF-QA is more than 2\%, around 1.5\% for both repeating action
and frame QA, and 0.08 MSE in case of repetition count. Although this
baseline produces great increments on the SVQA comparing to the experiment
Avg.Pool+MAC, the improvement on the TGIF-QA is minimal.

\textbf{CRN+MLP: }In order to evaluate how the reasoning module affects
the overall performance, we conduct this experiment by using a feed-forward
network as the reasoning module with the proposed visual representation
CRN. 

\textbf{CRN+MAC:} This is our proposed method in which we opt the
outputs of CRN for the knowledge base of MAC. We witness significant
improvements on all sub-tasks in the SVQA over the simplistic TRN
whilst results on the TGIF-QA dataset are less noticeable. The results
reveal the strong CRN's capability as well as a better suit of video
representation for reasoning over the TRN, especially in case of compositional
reasoning. The results also prove our earlier argument that sparsely
sampled frames from the video are insufficient to embrace fast-pace
actions/events such as repeating action and repetition count.

\begin{table*}
\caption{Comparison with the SOTA methods on SVQA. \label{tab:svqa}}

\centering{}%
\begin{tabular}{l|c|c|c|c|c|c|c|c|c|c|c|c|c|c|c|c}
\hline 
\multirow{2}{*}{} & \multirow{2}{*}{{\scriptsize{}Exist}} & \multirow{2}{*}{{\scriptsize{}Count}} & \multicolumn{3}{c|}{{\scriptsize{}Integer Comparison}} & \multicolumn{5}{c|}{{\scriptsize{}Attribute Comparison}} & \multicolumn{5}{c|}{{\scriptsize{}Query}} & \multirow{2}{*}{{\scriptsize{}All}}\tabularnewline
 &  &  & {\scriptsize{}More} & {\scriptsize{}Equal} & {\scriptsize{}Less} & {\scriptsize{}Color} & {\scriptsize{}Size} & {\scriptsize{}Type} & {\scriptsize{}Dir} & {\scriptsize{}Shape} & {\scriptsize{}Color} & {\scriptsize{}Size} & {\scriptsize{}Type} & {\scriptsize{}Dir} & {\scriptsize{}Shape} & \tabularnewline
\hline 
{\scriptsize{}SA(S) \cite{song2018explore}} & {\scriptsize{}51.7} & {\scriptsize{}36.3} & {\scriptsize{}72.7} & {\scriptsize{}54.8} & {\scriptsize{}58.6} & {\scriptsize{}52.2} & {\scriptsize{}53.6} & {\scriptsize{}52.7} & {\scriptsize{}53.0} & {\scriptsize{}52.3} & {\scriptsize{}29.0} & {\scriptsize{}54.0} & {\scriptsize{}55.7} & {\scriptsize{}38.1} & {\scriptsize{}46.3} & {\scriptsize{}43.1}\tabularnewline
{\scriptsize{}TA-GRU(T) \cite{song2018explore}} & {\scriptsize{}54.6} & {\scriptsize{}36.6} & {\scriptsize{}73.0} & {\scriptsize{}57.3} & {\scriptsize{}57.7} & {\scriptsize{}53.8} & {\scriptsize{}53.4} & {\scriptsize{}54.8} & {\scriptsize{}55.1} & {\scriptsize{}52.4} & {\scriptsize{}22.0} & {\scriptsize{}54.8} & {\scriptsize{}55.5} & {\scriptsize{}41.7} & {\scriptsize{}42.9} & {\scriptsize{}44.2}\tabularnewline
{\scriptsize{}SA+TA-GRU \cite{song2018explore}} & {\scriptsize{}52.0} & {\scriptsize{}38.2} & {\scriptsize{}74.3} & {\scriptsize{}57.7} & {\scriptsize{}61.6} & {\scriptsize{}56.0} & {\scriptsize{}55.9} & {\scriptsize{}53.4} & {\scriptsize{}57.5} & {\scriptsize{}53.0} & {\scriptsize{}23.4} & {\scriptsize{}63.3} & {\scriptsize{}62.9} & {\scriptsize{}43.2} & {\scriptsize{}41.7} & {\scriptsize{}44.9}\tabularnewline
\hline 
\textbf{\scriptsize{}CRN+MAC} & \textbf{\scriptsize{}72.8} & \textbf{\scriptsize{}56.7} & \textbf{\scriptsize{}84.5} & \textbf{\scriptsize{}71.7} & \textbf{\scriptsize{}75.9} & \textbf{\scriptsize{}70.5} & \textbf{\scriptsize{}76.2} & \textbf{\scriptsize{}90.7} & \textbf{\scriptsize{}75.9} & \textbf{\scriptsize{}57.2} & \textbf{\scriptsize{}76.1} & \textbf{\scriptsize{}92.8} & \textbf{\scriptsize{}91.0} & \textbf{\scriptsize{}87.4} & \textbf{\scriptsize{}85.4} & \textbf{\scriptsize{}75.8}\tabularnewline
\hline 
\end{tabular}
\end{table*}

\begin{table}
\caption{Comparison with the SOTA methods on TGIF-QA. For count, the lower
the better. R: ResNet, C: C3D features, F: flow features. \label{tab:tgif}}

\centering{}%
\begin{tabular}{lcccc}
\hline 
\noalign{\vskip0.5mm}
{\small{}Model} & {\small{}Action} & {\small{}Trans.} & {\small{}Frame} & {\small{}Count}\tabularnewline[0.5mm]
\hline 
\noalign{\vskip0.5mm}
{\small{}VIS+LSTM (aggr)\cite{ren2015exploring}} & {\small{}46.8} & {\small{}56.9} & {\small{}34.6} & {\small{}5.09}\tabularnewline[0.5mm]
\noalign{\vskip0.5mm}
{\small{}VIS+LSTM (avg)\cite{ren2015exploring}} & {\small{}48.8} & {\small{}34.8} & {\small{}35.0} & {\small{}4.80}\tabularnewline[0.5mm]
\noalign{\vskip0.5mm}
{\small{}VQA-MCB (aggr)\cite{fukui2016multimodal}} & {\small{}58.9} & {\small{}24.3} & {\small{}25.7} & {\small{}5.17}\tabularnewline[0.5mm]
\noalign{\vskip0.5mm}
{\small{}VQA-MCB (avg)\cite{fukui2016multimodal}} & {\small{}29.1} & {\small{}33.0} & {\small{}15.5} & {\small{}5.54}\tabularnewline[0.5mm]
\noalign{\vskip0.5mm}
{\small{}Yu et al.\cite{yu2017end}} & {\small{}56.1} & {\small{}64.0} & {\small{}39.6} & {\small{}5.13}\tabularnewline[0.5mm]
\noalign{\vskip0.5mm}
{\small{}ST(R+C)\cite{jang2017tgif}} & {\small{}60.1} & {\small{}65.7} & {\small{}48.2} & {\small{}4.38}\tabularnewline[0.5mm]
\noalign{\vskip0.5mm}
{\small{}ST-SP(R+C)\cite{jang2017tgif}} & {\small{}57.3} & {\small{}63.7} & {\small{}45.5} & {\small{}4.28}\tabularnewline[0.5mm]
\noalign{\vskip0.5mm}
{\small{}ST-SP-TP(R+C)\cite{jang2017tgif}} & {\small{}57.0} & {\small{}59.6} & {\small{}47.8} & {\small{}4.56}\tabularnewline[0.5mm]
\noalign{\vskip0.5mm}
{\small{}ST-TP(R+C)\cite{jang2017tgif}} & {\small{}60.8} & {\small{}67.1} & {\small{}49.3} & {\small{}4.40}\tabularnewline[0.5mm]
\noalign{\vskip0.5mm}
{\small{}ST-TP(R+F)\cite{jang2017tgif}} & {\small{}62.9} & {\small{}69.4} & {\small{}49.5} & {\small{}4.32}\tabularnewline[0.5mm]
\noalign{\vskip0.5mm}
{\small{}Co-memory(R+F)\cite{gao2018motion}} & {\small{}68.2} & {\small{}74.3} & {\small{}51.5} & {\small{}4.10}\tabularnewline[0.5mm]
\noalign{\vskip0.5mm}
{\small{}PSAC(R)\cite{li2019beyond}} & {\small{}70.4} & {\small{}76.9} & {\small{}55.7} & {\small{}4.27}\tabularnewline[0.5mm]
\noalign{\vskip0.5mm}
{\small{}HME(R+C)\cite{fan2019heterogeneous}} & \textbf{\small{}73.9} & {\small{}77.8} & {\small{}53.8} & \textbf{\small{}4.02}\tabularnewline[0.5mm]
\hline 
\noalign{\vskip0.5mm}
\textbf{\small{}CRN+MAC(R)} & {\small{}71.3} & \textbf{\small{}78.7} & \textbf{\small{}59.2} & {\small{}4.23}\tabularnewline[0.5mm]
\hline 
\end{tabular} 
\end{table}

\subsubsection{Benchmarking against SOTAs}

We also compare our proposed model with other state-of-the-art methods
on both datasets, as shown in Table~\ref{tab:svqa} (SVQA) and Table~\ref{tab:tgif}
(TGIF-QA). As the TGIF-QA is older, much effort has been spent on
benchmarking it and significant progress has been made in recent years.
The SVQA is new, and hence published results are not very indicative
of the latest advance in modeling.

For the SVQA, Table~\ref{tab:ablation1} and Table~\ref{tab:svqa}
reveal that the contributions of visual information to the overall
performance of the best known results are very little. This means
their system is largely suffered from the linguistic bias of the dataset
for the decision making process. In contrast, our proposed methods
do not seem to suffer from this problem. We establish new qualitatively
different SOTAs on all sub-tasks and a massive jump from 44.9\% accuracy
to 75.8\% accuracy overall. 

For the TGIF-QA dataset, Jang \emph{et al.} \cite{jang2017tgif} extended
winner models of the VQA 2016 challenge to evaluate on Video QA task,
namely VIS+LSTM \cite{ren2015exploring} and VQA-MCB \cite{fukui2016multimodal}.
Early fusion and late fusion are applied to both two approaches. We
also list here some other methods provided by \cite{jang2017tgif}
including those proposed in \cite{fukui2016multimodal} and \cite{yu2017end}.
Interestingly, none of the previous works reported ablation studies
of utilizing only textual cues as the input to assess the linguistic
bias of the dataset, and the fact that some of the reported methods
produced worse performance than this baseline. We suspect that the
improper integrating of visual information caused confusion to the
systems giving such low performance. In Table~\ref{tab:tgif}, SP
indicates spatial attention, ST presents temporal attention while
``R'', ``C'' and ``F'' indicate ResNet, C3D and optical-flow
features, respectively. Later, Gao \emph{et al.} \cite{gao2018motion}
greatly advanced the performance on this dataset with a co-memory
mechanism on two video feature streams. Li \emph{et al.} \cite{li2019beyond}
recently achieved respected performance on TGIF-QA with only ResNet
features by using a novel self-attention mechanism. Our method, which
is also relied on ResNet features only, could achieve new state-of-the-art
performance on the state transition task and the frame QA task with
a big gap comparing to prior works on the frame QA task. It appears
that methods using both appearance features (RestNet features) and
motion features (C3D or optical-flow features) perform bad on the
frame QA task, suggesting the need for an adaptive feature selection
mechanism. For action and counting tasks, although we have not outperformed
\cite{gao2018motion,fan2019heterogeneous}, it is not directly comparable
since they utilized motion in addition to appearance features. Our
method, on the other hand, models the temporal relationships without
explicitly using motion features and thus the action boundaries are
not clearly detected. We hypothesize that counting task needs a specific
network, as evident in recent work \cite{levy2015live,trott2017interpretable}.

\subsubsection{Qualitative Results}

Fig.~\ref{fig:qualcomp} shows example frames and associated question
types in the TGIF-QA and SVQA datasets. The figure also presents corresponding
responses by our proposed method, and those by ST-TP \cite{jang2017tgif}
(on the TGIF-QA) and TRN+MAC (our own special case of flat video representation,
on the SVQA) for reference. The questions clearly demonstrate challenges
that video QA systems must face such as visual ambiguity, subtlety,
compositional language understanding as well as concepts grounding.
The questions in the SVQA were designed for multi-step reasoning,
and the dual process system of CRN+MAC Net proves to be effective
in these cases.

%% file: discussion.tex
We have proposed a new differentiable architecture for learning to
reason in video question answering. The architecture is founded on
the premise that Video QA tasks necessitate a conditional dual process
of associative video cognition and deliberative multi-step reasoning,
given textual cues. The two processes are ordered in that the former
process prepares query-specific representation of video to support
the latter reasoning process. With that in mind, we designed a hierarchical
relational model for query-guided video representation named Clip-based
Relational Network (CRN) and integrated it with a generic neural reasoning
module (MAC Net) to infer an answer. The system is fully differentiable
and hence amenable to end-to-end training. Compared to existing state-of-the-arts
in Video QA, the new system is more modular, and thus open to accommodate
a wide range of low-level visual processing and high-level reasoning
capabilities. Tested on SVQA (synthetic) and TGIF-QA (real) datasets,
the proposed system demonstrates a new state-of-the-art performance
in a majority of cases. The gained margin is strongly evident in the
case where the system is defined for \textendash{} multi-step reasoning.

The proposed layered neural architecture is in line with proposals
in \cite{fodor1983modularity,harnad1990symbol}, where reactive perception
(System 1) precedes and is accessible to deliberative reasoning (System
2). Better perception capabilities will definitely make it easier
for visual reasoning. For example, action counting might benefit from
accurate explicit region proposals for objects and duration proposals
for action, rather than the implicit detection as currently implemented.
We also observed that the generic reasoning scheme of MAC net is surprisingly
powerful for the domain of Video QA, especially for the problems that
demand multi-step inference (e.g., on the SVQA dataset). This suggests
that it is worthy to spend effort to advance reasoning functionalities
for both general cases and in spatio-temporal settings. Finally, although
we have presented a seamless feedforward integration of System 1 and
System 2, it is still open on how the two systems interact.